\documentclass[onecolumn,journal]{IEEEtran} 
\usepackage{cite}
\usepackage{url}
\usepackage{graphicx}
\usepackage{epstopdf}
\usepackage{subfigure}
\usepackage{amsmath}
\usepackage{amsthm}
\usepackage{amsfonts}
\usepackage{amssymb}
\usepackage{latexsym}
\usepackage{moreverb} 
\usepackage{booktabs}
\usepackage{times}
\usepackage{indentfirst}
\usepackage{algorithmic}
\usepackage{algorithm}
\usepackage[table]{xcolor}  % for table with color
\usepackage{multirow}       % for table
\usepackage{subfigure}

\begin{document}
%\usepackage{svg}
%\usepackage{caption}
%\captionsetup[table]{position=below}
%=================================================================
%\firstpage{1} 
%\makeatletter 
%\setcounter{page}{\@firstpage} 
%\makeatother
%\pubvolume{xx}
%\issuenum{1}
%\articlenumber{1}
%\pubyear{2019}
%\copyrightyear{2019}
%\externaleditor{Academic Editor: name}
%\history{26 May 2019}

%=================================================================
% Full title of the paper (Capitalized)
\title{Accurate and Energy-Efficient Classification with Spiking Random Neural Network:\\
Corrected and Expanded Version}

% Authors, for the paper (add full first names)
\author{Khaled F. Hussain $^{1}$, Mohamed Yousef Bassyouni $^{1}$ and Erol Gelenbe $^{2}$\\
$^{1}$ Computer Science Department, Faculty of
Computers and Information, Assiut University, Assiut 71515, Egypt; khussain@aun.edu.eg; mohamed.mahdi@fci.au.edu.eg\\
$^{2}$ Institute of Theoretical \& Applied Informatics, Polish Academy of Sciences, ul. Baltycka 5, 44100 Gliwice, Poland,
\& Imperial College London, SW7 2BT United Kingdom; gelenbe.erol@gmail.com}

\maketitle
% Contact information of the corresponding author
%\corres{Correspondence: e-mail@e-mail.com; Tel.: +x-xxx-xxx-xxxx}

% Simple summary
%\simplesumm{}

% Abstract (Do not insert blank lines, i.e. \\) 
\begin{abstract}Artificial Neural Network (ANN) based techniques have dominated state-of-the-art results in most problems related to computer vision, audio recognition, and natural language processing in the past few years, resulting in strong industrial adoption from all leading technology companies worldwide. One of the major obstacles that have historically delayed large scale adoption of ANNs is the huge computational and power costs associated with training and testing (deploying) them. In the mean-time, Neuromorphic Computing platforms have recently achieved remarkable performance running more bio-realistic Spiking Neural Networks at high throughput and very low power consumption making them a natural alternative to ANNs. Here, we propose using the Random Neural Network (RNN), a spiking neural network with both theoretical and practical appealing properties, as a general purpose classifier that can match the classification power of ANNs on a number of tasks while enjoying all the features of a spiking neural network. This is demonstrated on a number of real-world classification datasets.\end{abstract}

% Keywords

%%%%%%%%%%%%%%%%%%%%%%%%%%%%%%%%%%%%%%%%%%

%%%%%%%%%%%%%%%%%%%%%%%%%%%%%%%%%%%%%%%%%%

\section{Introduction}

Despite being first proposed about 60 years ago \cite{rosenblatt1958perceptron}, only in the past few years have artificial neural networks (ANNs) become the de facto standard machine learning model \cite{lecun2015deep} achieving accurate state-of-the-art results for a wide range of problems ranging from image classification \cite{krizhevsky2012imagenet,he2016deep,szegedy2015going}, object detection \cite{girshick2014rich,redmon2016you}, semantic segmentation \cite{long2015fully,he2017mask}, face recognition \cite{schroff2015facenet,ranjan2017hyperface}, and text recognition \cite{graves2009novel,shi2017end}, to speech recognition \cite{hinton2012deep,graves2013speech,amodei2016deep}, natural language processing problems such as machine translation \cite{jean2014using,sutskever2014sequence}, language modeling \cite{bengio2003neural}, and question answering \cite{bordes2014question}. This has resulted in a huge industry-wide adoption from leading technology companies such as Google, Facebook, Microsoft, IBM, Yahoo!, Twitter, Adobe, and a quickly growing number of start-ups. 

One of the prominent reasons for this recent revival is that in order for ANNs to achieve such performance they need very large labeled datasets and huge computational power at a scale that only recently came into the hands of individual researchers in the form of GPUs \cite{raina2009large}, which kick-started the deep learning revolution in 2012 \cite{krizhevsky2012imagenet}. Since then, the trend for demanding more computation and more power consumption for such applications has largely increased.

Despite being initially bio-inspired architectures, ANNs have significant differences from actual biological neurons in how computations are performed by neurons, their structure (connection patterns and topologies of neurons), learning (how neurons adapt themselves to new observations), and communication (how inter-neuron data is encoded and passed). 

One of the main differences of ANNs compared to biological neurons, is how communication is done. While biological neurons use asynchronous trains of spikes in an  event-based, data-driven manner that adapts locally to its external stimulation pattern to communicate and encode data (though the specific encoding mechanism used by neurons is not totally understood), ANNs communicate in dense, continuous valued activations, which means that all ANN neurons are working at the same time, thus using lots of computation and energy to operate. 

Spiking neural networks leverage the benefit from biological neurons to communicate asynchronously in trains of spikes. Thus, spiking neural networks incorporate the concept of time, and instead of all neurons firing at the same time as the case with ANNs, in spiking neural networks neurons fire only when thier intrinsic potential (i.e. membrane voltage) reaches a specific threshold \cite{gerstner2014neuronal,hodgkin1952quantitative}. 

Neuroscientists have historically suggested several models for simulating how biological neurons communicate, and one of the simplest that is widely used is the integrate-and-fire (IF) model \cite{abbott1999lapicque}, in which the change in the membrane voltage $v_{mem}$ is given by the equation:
\begin{equation}
\frac{dv_{mem}(t)}{dt} = \sum_i \sum_{s\in S_i} w_i \delta (t-s)
\end{equation}
where $w_i$ is the weight of the $i$th incoming synapse, $\delta(.)$ is the Dirac delta function, and $S_i = {t^0_i, t^1_i, . . .}$ contains the spike times of the $i$th presynaptic neuron. If the membrane voltage crosses the spiking threshold $v_{thr}$, a spike is generated and the membrane voltage is reset to a reset potential $v_{res}$ \cite{diehl2015fast}. Several other models have also been proposed, such as te spike response model (SRM) \cite{jolivet2003spike}, and the Izhikevich model \cite{izhikevich2003simple}.

One of the prominent differences between spiking neural networks and ANNs is how they learn and adapt to new signals. ANNs have been predominantly trained in the literature using Backpropagation \cite{werbos1974beyond}, and with variants of stochastic gradient descent (SGD), which can summarised as moving the vector of network parameters or weights $\theta$ in the direction of the negative gradient of some loss function that characterizes the deviation network's current output from the ground truth labels of input data. 

Training spiking neural networks on the other hand is still an open research issue with many proposed solutions and no consensus \cite{tavanaei2018deep}. One of the most popular and biologically plausible learning methods in spiking neural networks is unsupervised learning using Spike Timing Dependent Plasticity (STDP) \cite{caporale2008spike,markram2011history}, in which the synaptic weight is adjusted in accordance with the relative spike times of the presynaptic and postsynaptic neurons. 

An important problem that has always faced using the popular gradient-based optimization algorithms in spiking neural networks is that both spike trains and the underlying membrane voltage are not differentiable at the time of spikes,. Thus researchers tried different approaches to alleviate this problem, and one of the most successful has been the workaround of first training an ANN and then converting it to a corresponding spiking neural network \cite{o2013real,diehl2015fast,neil2016learning}.

Though von Neumann architectures \cite{von1993first} work very well to run ANNs, it was suggested as early as the 1980s that they were not adequate for running the more realistic spiking neural networks models efficiently, and a new architecture was needed to realize thier power and computational efficiency \cite{zhao2010nanotube}. The recent success of ANNs has pushed this trend much faster. 

The main idea behind Neuromorphic Computing was to design Integrated Circuits (ICs) that are arranged and behave like living neurons (i.e. to mimic how the brain performs computation) \cite{mead1990neuromorphic}, and spiking neural network models of bioglogical neurons have historically been used as a guiding design in this process. After years of attempts, the past few years saw the demonstration of Neuromorphic Computing platforms with millions of neurons while requiring only milliWatts of power for thier operation such as TrueNorth \cite{merolla2014million}, SpiNNaker \cite{furber2014spinnaker}, and Loihi \cite{davies2018loihi}. A number of pattern classification applications were demonstrated to run efficiently and accurately on such chips while being orders of magnitude more efficient in terms of power consumption than an ANN on a von Neumann CPU or GPU running a similar task. 

The main source of this power saving is the asynchronous working and firing of spiking neural networks described earlier, so neurons fire and the chip consumes power only when needed, which is completely different than what happens in an ANN when all neurons are obliged to fire synchronously leading to unnecessary computation and energy consumption. This efficiency can be even increased when using input from neuromorphic sensors such as silicon retinas \cite{serrano2013128} or cochleas \cite{liu2010neuromorphic}, which create sparse, frame-free, and precisely timed trains of signals, with substantially reduced latencies compared to traditional frame-based approaches which produce large volumes of redundant data and therefore consume much power. This line of neuromorphic sensor design has been applied to vision sensors, auditory sensors, and olfactory sensors \cite{vanarse2016review}.

This paper presents results that show how a spiking model known as the Random Neural Network (RNN) \cite{RNN} can be used effectively and efficiently for Machine Learning applications, resulting in a level of performance comparable to the best ANN results. 

The rest of the paper is structured as follows: The Random Neural Network is described and reviewed in Section \ref{sec:rnn}; our experimental setup is described and results presented in Section \ref{sec:res}; the conclusions and future work are drawn in Section \ref{sec:cnc}

%%%%%%%%%%%%%%%%%%%%%%%%%%%%%%%%%%%%%%%%%%
\section{The Random Neural Network (RNN)}
\label{sec:rnn}

\begin{figure}[htbp]
  \centering
   \includegraphics[height=8cm,width=12cm]{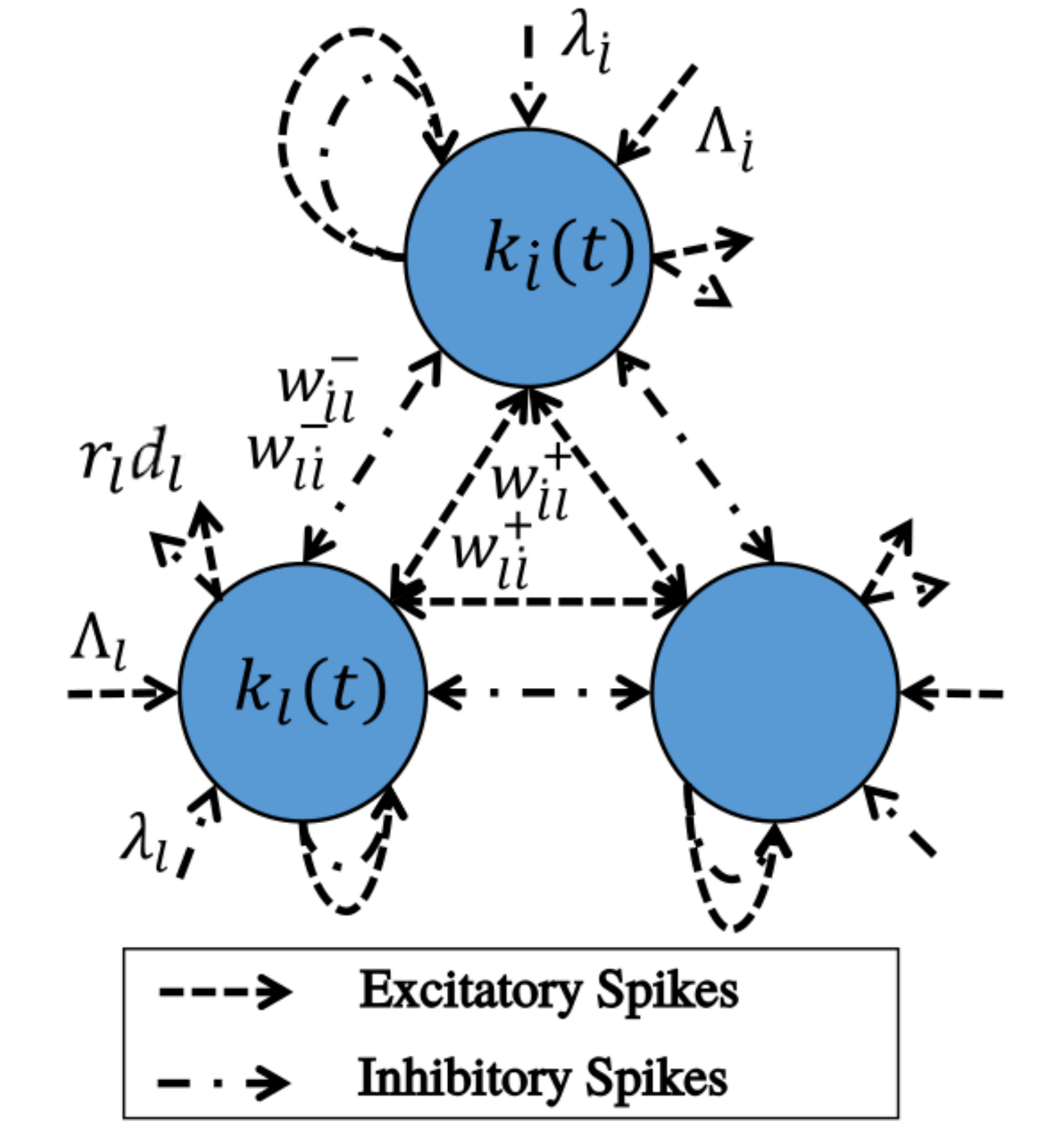}
   \centering
  \caption{Schematic representation of a Random Neural Network (RNN).}
  \label{fig:rnn}
\end{figure}

The Random Neural Network  (RNN) \cite{RNN,Stable,Stafy} has a powerful property of approximating continuous and bounded real-valued functions \cite{Approx}. This property serves as the foundation for RNN based learning algorithms, both for recurrent (containing feedback) and feedforward networks \cite{gelenbe1993learning}, and for Deep Learning \cite{Yin1,Yin2}.

The RNN has been used for modelling natural neuronal networks \cite{Biosystems}, and for protein alignment \cite{PhanSG12}.  It has been used in several image processing applications including the accurate evaluation of tumors from brain MRI scans \cite{MRI} and the compression of still and moving images \cite{Cramer1,Cramer2,hai2001video}. It was recently introduced as a tool for predicting the toxicity of chemical compounds \cite{ICANN18}.

In the field of computer networks, the RNN has been used to build distributed controllers for quality of service routing in packet networks \cite{CPN,Zarina,Brun} and in the design of Software Defined Network controllers for the Internet \cite{Francois1,Francois2}. Real-time optimised task allocation algorithms in Cloud systems \cite{Wang2} have also been built and tested. Recent applications has addressed the use of the RNN to detect network attacks \cite{Worms} and attacks on Internet of Things (IoT) gateways \cite{IoT}.

On the other hand, G-Networks \cite{GN1} are a family of queueing networks that generalize the RNN model, and like the RNN they have a convenient and computationally efficient ``product form ''mathematical solution. The computation of the state of a G-Network is obtained via a simple fixed-point iteration, and the existence and uniqueness of the solution to the key G-Network state equation is shown in \cite{GN3}. G-Networks incorporate useful primitives, such as the transfer of jobs between servers or the removal of batches of jobs from excessively busy servers, that were developed in a series of successive papers \cite{GN2,GN4,GN5,GN6}. They have a wealth of diverse applications as a tool to analyse and optimise the effects of dynamic load balancing in large scale networks and distributed computer systems \cite{Morfo}. They are also used to model Gene Regulatory Networks \cite{GelenbeRegNets2007,KimG12}. A recent application of G-Networks is to the modelling of systems which operate with intermittent sources of energy, known as Energy Packet Networks \cite{EPN,Marin,Fourneau,Ceran2,NOLTA}.

\subsection{The RNN as a Mathematical Model}

Figure~\ref{fig:rnn} gives a schematic diagram of a RNN for a system with $L$ neurons. The state of neuron $i$ at time $t$ is represented by a non-negative integer called its potential, denoted by $k_i(t) \geq 0$. Network state at time $t$ is a vector
$k(t) = (k_1(t), ... , k_i(t), \dotsc, k_L (t))$. When an excitation spike arrives to neuron $l$, the state of neuron $i$ is changed from $k_i(t)$ to $k_i(t)+1$ . When an inhibition spike arrives to neuron $i$, its state from $k_i(t)$ to $k_i(t)-1$ if $k_i(t)>0$, and does not change if $k_i(t)=0$.  

Each neuron $i$ receives exogenous or external excitation spikes in the form of two independent Poisson processes of rates $\Lambda_i\geq 0$ and $\lambda_i\geq 0$, respectively. These represent the external inputs to the network.

Neuron $i$ can emit a spike if $k_i(t)$ is positive (i.e. it is excited); this occurs with probability $r_i\Delta(t)+ o(\Delta(t)$, where $r_i \geq 0$ is the ``firing rate" of neuron $i$. Ehen this happens, neuron $i$'s state changes from $k_i(t)$ to $k_i(t)-1$. The spikes are then sent .

Spikes are sent from the outside world to neuron $i$ as a positive signal according to Poisson processes of rate $\Lambda_l$ or as a negative signal according also to Poisson processes of rate $\lambda_l$

Each spike is sent out from neuron $i$ to neuron $j$ as a positive or excitatory spike with probability $p^+_{ij}$, or as a negative or inhibitory spike spike with probability $p^-_{ij}$ , or it departs from the network with probability $d_i$
which represents the probability that spikes emanating from a neuron may be ``lost'' for a variety of reasons, and does not reach any target neuron $j$. The sum of these probabilities must be one:
\begin{equation}
d_i + \sum_{j=1}^L[p^+_{ij} + p^-_{ij}] = 1, \forall i~.
\label{eq:e1}
\end{equation}
The spikes are sent out from neuron $i$ to neuron $j$ at rates:
\begin{equation}
w^+_{ij} = r_i p^+_{ij} \geq 0,~w^-_{ij} = r_li\: p^-_{ij} \geq 0,
\label{eq:e3}
\end{equation}
and $w^+_{ij}$ and $w^-_{ij}$ are called the excitatory and inhibitory weights, respectively.

Combining equations (\ref{eq:e1}) and (\ref{eq:e3}) we get:
\begin{equation}
r_i= \frac{\sum_{j=1}^L[ w^+_{ij} + w^-_{ij}]}{1-d_i}~.
\end{equation}
Let $q_i = \lim_{t \to \infty} Prob(k_lit) > 0)$ denote the stationary excitation probability of neuron $i$. The total arrival rates of positive signals $\Omega_i$ and negative signals $\Omega^-_i$ , for $i = 1, \dotsc, L$, can be calculated from the following nonlinear system of equations:
\begin{equation}
\Omega^+_i= \Lambda_i + \sum_{j=1}^L q_i w^+_{ij},~\Omega^-_i = \lambda_i + \sum_{j=1}^L q_i w^-_{ij}
\end{equation}
It has been proven that the $q_i$ can be directly calculated by the following system of equations:
\begin{equation}
q_i= \min \left \{ 1, \frac{\Omega^+_i}{r_i+\Omega^-_i} \right \},~1\leq i\leq L,
\label{eq:4}
\end{equation}
and furthermore that if all the resulting $q_i$ are strictly less than one, then:
\begin{equation}
Prob[k_1(t)=k_1,~...~,k_L(t)=k_L]=\Pi_{i=1}^L q_i^{k_i}(1-q_i),
\end{equation}
which is known as a ``Product Form Solution''.

The existence of a solution to the system of $L$ non-linear equations (\ref{eq:4}) and the uniqueness of the solution has been proven \cite{gelenbe1993learning}. Therefore, the states of the RNN can be obtained by solving  (\ref{eq:4}), for instance using a fixed-point iteration.

%%%%%%%%%%%%%%%%%%%%%%%%%%%%%%%%%%%%%%%%%%
\section{Experimental Results using the RNN}
\label{sec:res}

We know that spiking neural network-based RNNs are empirically at least as powerful as ANNs in this category of classification problems. It was already shown that RNNs are universal function approximators for continuous and bounded functions \cite{gelenbe2006function,gelenbe1999function}, and hence as computationally capable as multi-layer perceptrons in this respect \cite{hornik1989multilayer}. 
However, RNNs have been shown to have an unique solution even in the recurrent case \cite{gelenbe1993learning,Multiclass,hai2001video}, and furthermore in the recurrent case the RNN's learning algorithm is still of low polynomial time and space complexity \cite{gelenbe2001learning,Laser,Simulation}.

Here we will provide experimental results on a number of benchmark real world classification datasets that are widely used in the literature. These experimental results can be reproduced using the software and and data stored at the web site \url{www.github.com/ASDen/Random_Neural_Network}. 

Another web site that specializes in \underline{Deep Learning with the Random Neural Network}  can be found at \url{https://github.com/yinyongh/DenseRandomNet}.

\subsection{Evaluation Setup}

\begin{table}[h!]
\centering
\begin{tabular}{ c|c|c|c } 
 \hline
 \emph{Dataset} & \emph{\# Attributes} & \emph{\# Features} & \emph{\# Output Classes}  \\ \hline
 Iris & 4 & 150 & 3 \\ 
 Breast Cancer Wisconsin  & 9 & 699 & 2 \\ 
 Glass & 9 & 214 & 7 \\
 Ovarian cancer & 100 & 216 & 2 \\
 \hline
\end{tabular}
\caption{Names and statistics of the Datasets that are used.}
\label{tbl:dsets}
\end{table}

Table \ref{tbl:dsets} shows statistics about the used datasets. We use the Iris, Breast Cancer, and Glass datasets from the UCI machine learning repository \cite{lichman2013uci} and the Ovarian cancer dataset \cite{petricoin2002use}. We train the RNN using the procedure described in \cite{hussain2016road} based on \cite{Learning93}; the related software
and data sets are at \url{www.github.com/ASDen/Random_Neural_Network}. The algorithm can be summarized as follows: 
\begin{enumerate}[leftmargin=*,labelsep=4.9mm]
\item Assume the given dataset has $K$ pairs of input training patterns for an $L-vector$  $x_k=(x_{1l},~...~,x_{Lk})$ associated with the output value $L-vector$ $y_k=(y_{1l},~...~,y_{Lk})$.
\item	Initialize the weights $w^+_{ij}$ and $w^-_{ij}$ , $\forall ( i,~j)$, to random values between zero and one.
\item	Set the external inhibitory arrival rates to zero.
\item	Set the external excitatory input rates for the input neurons, $\Lambda_k = x_k$, where $x_k$ is the $k$th input training pattern.
\item   Solve the nonlinear system of equations (\ref{eq:4}) to obtain each neuron's stationary excitation probability $q_i, \forall i$~.
\item   For all the given $(input,~output)$ values $(x_k,~y_k)$, iterate through the RNN recurrent network learning algorithm \cite{Learning93,Learning02} till convergence, updating at each step the weights $w^+_{ij}$ and $w^-_{ij}$ , $\forall (i,~j)$, so as to minimize the following error function:
\begin{equation}
E = \sum_{k=1}^K \sum_{i=1}^L [q_{ik} - y_{ik}]^2
\end{equation}
\end{enumerate}

\subsection{Dataset Description}
Here we give a brief description of the four datasets used for evaluation:
\begin{enumerate}[leftmargin=*,labelsep=4.9mm]
\item	Iris dataset \cite{lichman2013uci}: Each instance is described by four plants attributes (sepal length and width, and petal length and width) all are real numbers and the task is to recognize which class of Iris plants (Iris Setosa, Iris Versicolour, or Iris Virginica) a given test instance belongs to.

\item	Breast Cancer Wisconsin dataset \cite{lichman2013uci}: Each instance is described by 9 numerical attributes, that range from 1 to 10. The attributes include the clump thickness, uniformity of cell size, uniformity of cell shape, marginal adhesion, single epithelial cell size, bare nuclei, bland chromatin, normal nucleoli and mitoses. This breast cancer databases was obtained from the University of Wisconsin Hospitals, Madison from Dr. William H. Wolberg. The task is to recognize the class of the beast cancer (benign or malignant).

\item	Glass dataset \cite{lichman2013uci}: Each instance has 9 continuous attributes, including the refractive index and the unit measurements of sodium, magnesium, aluminum, silicon, potassium, calcium, barium and iron. There are in total 7 types of glass, while there are instances of only 6 types of glass in the dataset. The task is to using the 9 attributes to recognize which type of glass this instance belongs to (whether it is windows glass or non-window glass).

\item	Ovarian Cancer dataset \cite{petricoin2002use}: From the FDA-NCI Clinical Proteomics Program Databank, the dataset comprises 216 patients, out of which 121 are ovarian cancer patients and 95 are normal patients. Each instance has 100 attributes, each of which represents the ion intensity level at a specific mass-charge value . The task is to recognize the class of the ovarian cancer (benign or malignant).
\end{enumerate}

\subsection{Results}

Tables \ref{tbl:iris}, \ref{tbl:bc}, \ref{tbl:gls}, and \ref{tbl:ovc} display confusion matrices of RNN on the four datasets. Table \ref{tbl:dsets} compares the accuracy of RNNs against ANNs some UCI datasets. For the ANN results, we use results obtained on UCI datasets from the comprehensive study in \cite{fernandez2014we}. We can clearly see in \ref{tbl:dsets} that RNNs are at least as powerful as ANNs in these datasets and can deliver excellent classification accuracy.

\begin{table}[h!]
\centering
\begin{tabular}{ c c c c } 
 \hline
 \emph{Class} & \emph{Setosa} & \emph{Versicolour} & \emph{Virginica}  \\ \hline
 \emph{Setosa} & 1.0 & 0.0 & 0.0 \\ 
 \emph{Versicolour} & 0.0 & 1.0 & 0.0 \\ 
 \emph{Virginica} & 0.0 & 0.0 & 1.0 \\
 \hline
\end{tabular}
\caption{Confusion matrix for Iris dataset.}
\label{tbl:iris}
\end{table}

\begin{table}[h!]
\centering
\begin{tabular}{ c c c } 
 \hline
 \emph{Class} & \emph{Positive} & \emph{Negative} \\ \hline
 \emph{Postivie} & 0.984 & 0.016 \\ 
 \emph{Negative} & 0.067 & 0.933 \\ 
 \hline
\end{tabular}
\caption{Confusion matrix for Breast Cancer dataset.}
\label{tbl:bc}
\end{table}

\begin{table}[h!]
\centering
\begin{tabular}{ c c c } 
 \hline
 \emph{Class} & \emph{Positive} & \emph{Negative} \\ \hline
 \emph{Postivie} & 1.0 & 0.0 \\ 
 \emph{Negative} & 0.127 & 0.8139 \\ 
 \hline
\end{tabular}
\caption{Confusion matrix for Glass dataset.}
\label{tbl:gls}
\end{table}

\begin{table}[h!]
\centering
\begin{tabular}{ c c c } 
 \hline
 \emph{Class} & \emph{Postive} & \emph{Negative} \\ \hline
 \emph{Postive} & 1.0 & 0.0 \\ 
 \emph{Negative} & 0.1064 & 0.8936 \\ 
 \hline
\end{tabular}
\caption{Confusion matrix for Ovarian Cancer dataset.}
\label{tbl:ovc}
\end{table}

\begin{table}[h!]
\centering
\begin{tabular}{ c|c|c } 
 \hline
 \emph{Dataset} & \emph{RNN} & \emph{ANN}  \\ \hline
 Iris & \textbf{1.0} & 0.959  \\ 
 Breast Cancer Wisconsin  & \textbf{0.964} & 0.963 \\ 
% Glass & 0.58 & 0.59 & \textbf{0.785} \\
 \hline
\end{tabular}
\caption{Accuracy comparison between RNN, and ANN. Best result in each dataset is in bold.}
\label{tbl:dsets}
\end{table}

%%%%%%%%%%%%%%%%%%%%%%%%%%%%%%%%%%%%%%%%%%
\section{Conclusions}
\label{sec:cnc}

In this paper, we have motivated the need for power and computation efficiency in neuromorphic models
that are used in Machine Learning, and discussed some of the shortcomings of current ANN models in this respect. We have briefly presented and reviewed some spiking neural networkmodels used for neuromorphic computing as a possible alternative due to their low power usage and computational efficiency, but we have also
indicated their limitations with regard to being harder to train and their generalization performance.

We have then presented a specific spiking neural network model, the Random Neural Network (RNN)  that was first introduced in \cite{gelenbe1989random}. We have indicated that the RNN's special analytical properties makes it much easier to train, and we have shown empirically that it provides a generalization performance that is at least as powerful as conventional ANNs for a number of real world classification datasets, while achieving the efficiency associated with spiking neural networks.

\bibliographystyle{IEEEtran}
\bibliography{sbib}

\begin{thebibliography}{10}
\providecommand{\url}[1]{#1}
\csname url@samestyle\endcsname
\providecommand{\newblock}{\relax}
\providecommand{\bibinfo}[2]{#2}
\providecommand{\BIBentrySTDinterwordspacing}{\spaceskip=0pt\relax}
\providecommand{\BIBentryALTinterwordstretchfactor}{4}
\providecommand{\BIBentryALTinterwordspacing}{\spaceskip=\fontdimen2\font plus
\BIBentryALTinterwordstretchfactor\fontdimen3\font minus
  \fontdimen4\font\relax}
\providecommand{\BIBforeignlanguage}[2]{{%
\expandafter\ifx\csname l@#1\endcsname\relax
\typeout{** WARNING: IEEEtran.bst: No hyphenation pattern has been}%
\typeout{** loaded for the language `#1'. Using the pattern for}%
\typeout{** the default language instead.}%
\else
\language=\csname l@#1\endcsname
\fi
#2}}
\providecommand{\BIBdecl}{\relax}
\BIBdecl

\bibitem{rosenblatt1958perceptron}
F.~Rosenblatt, ``The perceptron: a probabilistic model for information storage
  and organization in the brain.'' \emph{Psychological review}, vol.~65, no.~6,
  p. 386, 1958.

\bibitem{lecun2015deep}
Y.~LeCun, Y.~Bengio, and G.~Hinton, ``Deep learning,'' \emph{nature}, vol. 521,
  no. 7553, p. 436, 2015.

\bibitem{krizhevsky2012imagenet}
A.~Krizhevsky, I.~Sutskever, and G.~E. Hinton, ``Imagenet classification with
  deep convolutional neural networks,'' in \emph{Advances in neural information
  processing systems}, 2012, pp. 1097--1105.

\bibitem{he2016deep}
K.~He, X.~Zhang, S.~Ren, and J.~Sun, ``Deep residual learning for image
  recognition,'' in \emph{Proceedings of the IEEE conference on computer vision
  and pattern recognition}, 2016, pp. 770--778.

\bibitem{szegedy2015going}
C.~Szegedy, W.~Liu, Y.~Jia, P.~Sermanet, S.~Reed, D.~Anguelov, D.~Erhan,
  V.~Vanhoucke, and A.~Rabinovich, ``Going deeper with convolutions,'' in
  \emph{Proceedings of the IEEE conference on computer vision and pattern
  recognition}, 2015, pp. 1--9.

\bibitem{girshick2014rich}
R.~Girshick, J.~Donahue, T.~Darrell, and J.~Malik, ``Rich feature hierarchies
  for accurate object detection and semantic segmentation,'' in
  \emph{Proceedings of the IEEE conference on computer vision and pattern
  recognition}, 2014, pp. 580--587.

\bibitem{redmon2016you}
J.~Redmon, S.~Divvala, R.~Girshick, and A.~Farhadi, ``You only look once:
  Unified, real-time object detection,'' in \emph{Proceedings of the IEEE
  conference on computer vision and pattern recognition}, 2016, pp. 779--788.

\bibitem{long2015fully}
J.~Long, E.~Shelhamer, and T.~Darrell, ``Fully convolutional networks for
  semantic segmentation,'' in \emph{Proceedings of the IEEE conference on
  computer vision and pattern recognition}, 2015, pp. 3431--3440.

\bibitem{he2017mask}
K.~He, G.~Gkioxari, P.~Doll{\'a}r, and R.~Girshick, ``Mask r-cnn,'' in
  \emph{Computer Vision (ICCV), 2017 IEEE International Conference on}.\hskip
  1em plus 0.5em minus 0.4em\relax IEEE, 2017, pp. 2980--2988.

\bibitem{schroff2015facenet}
F.~Schroff, D.~Kalenichenko, and J.~Philbin, ``Facenet: A unified embedding for
  face recognition and clustering,'' in \emph{Proceedings of the IEEE
  conference on computer vision and pattern recognition}, 2015, pp. 815--823.

\bibitem{ranjan2017hyperface}
R.~Ranjan, V.~M. Patel, and R.~Chellappa, ``Hyperface: A deep multi-task
  learning framework for face detection, landmark localization, pose
  estimation, and gender recognition,'' \emph{IEEE Transactions on Pattern
  Analysis and Machine Intelligence}, 2017.

\bibitem{graves2009novel}
A.~Graves, M.~Liwicki, S.~Fern{\'a}ndez, R.~Bertolami, H.~Bunke, and
  J.~Schmidhuber, ``A novel connectionist system for unconstrained handwriting
  recognition,'' \emph{IEEE transactions on pattern analysis and machine
  intelligence}, vol.~31, no.~5, pp. 855--868, 2009.

\bibitem{shi2017end}
B.~Shi, X.~Bai, and C.~Yao, ``An end-to-end trainable neural network for
  image-based sequence recognition and its application to scene text
  recognition,'' \emph{IEEE transactions on pattern analysis and machine
  intelligence}, vol.~39, no.~11, pp. 2298--2304, 2017.

\bibitem{hinton2012deep}
G.~Hinton, L.~Deng, D.~Yu, G.~E. Dahl, A.-r. Mohamed, N.~Jaitly, A.~Senior,
  V.~Vanhoucke, P.~Nguyen, T.~N. Sainath \emph{et~al.}, ``Deep neural networks
  for acoustic modeling in speech recognition: The shared views of four
  research groups,'' \emph{IEEE Signal processing magazine}, vol.~29, no.~6,
  pp. 82--97, 2012.

\bibitem{graves2013speech}
A.~Graves, A.-r. Mohamed, and G.~Hinton, ``Speech recognition with deep
  recurrent neural networks,'' in \emph{Acoustics, speech and signal processing
  (icassp), 2013 ieee international conference on}.\hskip 1em plus 0.5em minus
  0.4em\relax IEEE, 2013, pp. 6645--6649.

\bibitem{amodei2016deep}
D.~Amodei, S.~Ananthanarayanan, R.~Anubhai, J.~Bai, E.~Battenberg, C.~Case,
  J.~Casper, B.~Catanzaro, Q.~Cheng, G.~Chen \emph{et~al.}, ``Deep speech 2:
  End-to-end speech recognition in english and mandarin,'' in
  \emph{International Conference on Machine Learning}, 2016, pp. 173--182.

\bibitem{jean2014using}
S.~Jean, K.~Cho, R.~Memisevic, and Y.~Bengio, ``On using very large target
  vocabulary for neural machine translation,'' \emph{arXiv preprint
  arXiv:1412.2007}, 2014.

\bibitem{sutskever2014sequence}
I.~Sutskever, O.~Vinyals, and Q.~V. Le, ``Sequence to sequence learning with
  neural networks,'' in \emph{Advances in neural information processing
  systems}, 2014, pp. 3104--3112.

\bibitem{bengio2003neural}
Y.~Bengio, R.~Ducharme, P.~Vincent, and C.~Jauvin, ``A neural probabilistic
  language model,'' \emph{Journal of machine learning research}, vol.~3, no.
  Feb, pp. 1137--1155, 2003.

\bibitem{bordes2014question}
A.~Bordes, S.~Chopra, and J.~Weston, ``Question answering with subgraph
  embeddings,'' \emph{arXiv preprint arXiv:1406.3676}, 2014.

\bibitem{raina2009large}
R.~Raina, A.~Madhavan, and A.~Y. Ng, ``Large-scale deep unsupervised learning
  using graphics processors,'' in \emph{Proceedings of the 26th annual
  international conference on machine learning}.\hskip 1em plus 0.5em minus
  0.4em\relax ACM, 2009, pp. 873--880.

\bibitem{gerstner2014neuronal}
W.~Gerstner, W.~M. Kistler, R.~Naud, and L.~Paninski, \emph{Neuronal dynamics:
  From single neurons to networks and models of cognition}.\hskip 1em plus
  0.5em minus 0.4em\relax Cambridge University Press, 2014.

\bibitem{hodgkin1952quantitative}
A.~L. Hodgkin and A.~F. Huxley, ``A quantitative description of membrane
  current and its application to conduction and excitation in nerve,''
  \emph{The Journal of physiology}, vol. 117, no.~4, pp. 500--544, 1952.

\bibitem{abbott1999lapicque}
L.~F. Abbott, ``Lapicque’s introduction of the integrate-and-fire model
  neuron (1907),'' \emph{Brain research bulletin}, vol.~50, no. 5-6, pp.
  303--304, 1999.

\bibitem{diehl2015fast}
P.~U. Diehl, D.~Neil, J.~Binas, M.~Cook, S.-C. Liu, and M.~Pfeiffer,
  ``Fast-classifying, high-accuracy spiking deep networks through weight and
  threshold balancing,'' in \emph{Neural Networks (IJCNN), 2015 International
  Joint Conference on}.\hskip 1em plus 0.5em minus 0.4em\relax IEEE, 2015, pp.
  1--8.

\bibitem{jolivet2003spike}
R.~Jolivet, J.~Timothy, and W.~Gerstner, ``The spike response model: a
  framework to predict neuronal spike trains,'' in \emph{Artificial Neural
  Networks and Neural Information Processing—ICANN/ICONIP 2003}.\hskip 1em
  plus 0.5em minus 0.4em\relax Springer, 2003, pp. 846--853.

\bibitem{izhikevich2003simple}
E.~M. Izhikevich, ``Simple model of spiking neurons,'' \emph{IEEE Transactions
  on neural networks}, vol.~14, no.~6, pp. 1569--1572, 2003.

\bibitem{werbos1974beyond}
P.~Werbos, ``Beyond regression: New tools for prediction and analysis in the
  behavioral sciences,'' \emph{Ph. D. dissertation, Harvard University}, 1974.

\bibitem{tavanaei2018deep}
A.~Tavanaei, M.~Ghodrati, S.~R. Kheradpisheh, T.~Masquelier, and A.~S. Maida,
  ``Deep learning in spiking neural networks,'' \emph{arXiv preprint
  arXiv:1804.08150}, 2018.

\bibitem{caporale2008spike}
N.~Caporale and Y.~Dan, ``Spike timing--dependent plasticity: a hebbian
  learning rule,'' \emph{Annu. Rev. Neurosci.}, vol.~31, pp. 25--46, 2008.

\bibitem{markram2011history}
H.~Markram, W.~Gerstner, and P.~J. Sj{\"o}str{\"o}m, ``A history of
  spike-timing-dependent plasticity,'' \emph{Frontiers in synaptic
  neuroscience}, vol.~3, p.~4, 2011.

\bibitem{o2013real}
P.~O'Connor, D.~Neil, S.-C. Liu, T.~Delbruck, and M.~Pfeiffer, ``Real-time
  classification and sensor fusion with a spiking deep belief network,''
  \emph{Frontiers in neuroscience}, vol.~7, p. 178, 2013.

\bibitem{neil2016learning}
D.~Neil, M.~Pfeiffer, and S.-C. Liu, ``Learning to be efficient: algorithms for
  training low-latency, low-compute deep spiking neural networks,'' in
  \emph{Proceedings of the 31st Annual ACM Symposium on Applied
  Computing}.\hskip 1em plus 0.5em minus 0.4em\relax ACM, 2016, pp. 293--298.

\bibitem{von1993first}
J.~Von~Neumann, ``First draft of a report on the edvac,'' \emph{IEEE Annals of
  the History of Computing}, vol.~15, no.~4, pp. 27--75, 1993.

\bibitem{zhao2010nanotube}
W.~Zhao, G.~Agnus, V.~Derycke, A.~Filoramo, J.~Bourgoin, and C.~Gamrat,
  ``Nanotube devices based crossbar architecture: toward neuromorphic
  computing,'' \emph{Nanotechnology}, vol.~21, no.~17, p. 175202, 2010.

\bibitem{mead1990neuromorphic}
C.~Mead, ``Neuromorphic electronic systems,'' \emph{Proceedings of the IEEE},
  vol.~78, no.~10, pp. 1629--1636, 1990.

\bibitem{merolla2014million}
P.~A. Merolla, J.~V. Arthur, R.~Alvarez-Icaza, A.~S. Cassidy, J.~Sawada,
  F.~Akopyan, B.~L. Jackson, N.~Imam, C.~Guo, Y.~Nakamura \emph{et~al.}, ``A
  million spiking-neuron integrated circuit with a scalable communication
  network and interface,'' \emph{Science}, vol. 345, no. 6197, pp. 668--673,
  2014.

\bibitem{furber2014spinnaker}
S.~B. Furber, F.~Galluppi, S.~Temple, and L.~A. Plana, ``The spinnaker
  project,'' \emph{Proceedings of the IEEE}, vol. 102, no.~5, pp. 652--665,
  2014.

\bibitem{davies2018loihi}
M.~Davies, N.~Srinivasa, T.-H. Lin, G.~Chinya, Y.~Cao, S.~H. Choday, G.~Dimou,
  P.~Joshi, N.~Imam, S.~Jain \emph{et~al.}, ``Loihi: A neuromorphic manycore
  processor with on-chip learning,'' \emph{IEEE Micro}, vol.~38, no.~1, pp.
  82--99, 2018.

\bibitem{serrano2013128}
T.~Serrano-Gotarredona and B.~Linares-Barranco, ``A 128$\times$ 128 1.5\%
  contrast sensitivity 0.9\% fpn 3 $\mu$s latency 4 mw asynchronous frame-free
  dynamic vision sensor using transimpedance preamplifiers.'' \emph{J.
  Solid-State Circuits}, vol.~48, no.~3, pp. 827--838, 2013.

\bibitem{liu2010neuromorphic}
S.-C. Liu and T.~Delbruck, ``Neuromorphic sensory systems,'' \emph{Current
  opinion in neurobiology}, vol.~20, no.~3, pp. 288--295, 2010.

\bibitem{vanarse2016review}
A.~Vanarse, A.~Osseiran, and A.~Rassau, ``A review of current neuromorphic
  approaches for vision, auditory, and olfactory sensors,'' \emph{Frontiers in
  neuroscience}, vol.~10, p. 115, 2016.

\bibitem{RNN}
E.~Gelenbe, ``Random neural networks with negative and positive signals and
  product form solution,'' \emph{Neural Computation}, vol.~1, pp. 502--510,
  1989.

\bibitem{Stable}
------, ``R{\'e}seaux neuronaux al{\'e}atoires stables,'' \emph{Comptes rendus
  de l'Acad{\'e}mie des Sciences. S{\'e}rie 2, M{\'e}canique, Physique, Chimie,
  Sciences de l'Univers, Sciences de la Terre}, vol. 310, no.~3, pp. 177--180,
  1990.

\bibitem{Stafy}
E.~Gelenbe and A.~Stafylopatis, ``Global behavior of homogeneous random neural
  systems,'' \emph{Applied mathematical modelling}, vol.~15, no.~10, pp.
  534--541, 1991.

\bibitem{Approx}
\BIBentryALTinterwordspacing
E.~Gelenbe, Z.~Mao, and Y.~Li, ``Function approximation with spiked random
  networks,'' \emph{{IEEE} Trans. Neural Networks}, vol.~10, no.~1, pp. 3--9,
  1999. [Online]. Available: \url{https://doi.org/10.1109/72.737488}
\BIBentrySTDinterwordspacing

\bibitem{gelenbe1993learning}
E.~Gelenbe, ``Learning in the recurrent random neural network,'' \emph{Neural
  computation}, vol.~5, no.~1, pp. 154--164, 1993.

\bibitem{Yin1}
E.~Gelenbe and Y.~Yin, ``Deep learning with random neural networks,'' in
  \emph{Neural Networks (IJCNN), 2016 International Joint Conference on}.\hskip
  1em plus 0.5em minus 0.4em\relax IEEE, 2016, pp. 1633--1638.

\bibitem{Yin2}
Y.~Yin and E.~Gelenbe, ``Single-cell based random neural network for deep
  learning,'' in \emph{Neural Networks (IJCNN), 2017 International Joint
  Conference on}.\hskip 1em plus 0.5em minus 0.4em\relax IEEE, 2017, pp.
  86--93.

\bibitem{Biosystems}
E.~Gelenbe and C.~Cramer, ``Oscillatory corticothalamic response to
  somatosensory input,'' \emph{Biosystems}, vol.~48, no. 1-3, pp. 67--75, 1998.

\bibitem{PhanSG12}
\BIBentryALTinterwordspacing
H.~T.~T. Phan, M.~J.~E. Sternberg, and E.~Gelenbe, ``Aligning protein-protein
  interaction networks using random neural networks,'' in \emph{2012 {IEEE}
  International Conference on Bioinformatics and Biomedicine, {BIBM} 2012,
  Philadelphia, PA, USA, October 4-7, 2012}, 2012, pp. 1--6. [Online].
  Available: \url{https://doi.org/10.1109/BIBM.2012.6392664}
\BIBentrySTDinterwordspacing

\bibitem{MRI}
E.~Gelenbe, Y.~Feng, and K.~R.~R. Krishnan, ``Neural network methods for
  volumetric magnetic resonance imaging of the human brain,'' \emph{Proceedings
  of the IEEE}, vol.~84, no.~10, pp. 1488--1496, 1996.

\bibitem{Cramer1}
\BIBentryALTinterwordspacing
E.~Gelenbe, M.~Sungur, C.~Cramer, and P.~Gelenbe, ``Traffic and video quality
  with adaptive neural compression,'' \emph{Multimedia Syst.}, vol.~4, no.~6,
  pp. 357--369, 1996. [Online]. Available:
  \url{https://doi.org/10.1007/s005300050037}
\BIBentrySTDinterwordspacing

\bibitem{Cramer2}
\BIBentryALTinterwordspacing
C.~Cramer and E.~Gelenbe, ``Video quality and traffic qos in learning-based
  subsampled and receiver-interpolated video sequences,'' \emph{{IEEE} Journal
  on Selected Areas in Communications}, vol.~18, no.~2, pp. 150--167, 2000.
  [Online]. Available: \url{https://doi.org/10.1109/49.824788}
\BIBentrySTDinterwordspacing

\bibitem{hai2001video}
F.~Hai, K.~F. Hussain, E.~Gelenbe, and R.~K. Guha, ``Video compression with
  wavelets and random neural network approximations,'' in \emph{Applications of
  Artificial Neural Networks in Image Processing VI}, vol. 4305.\hskip 1em plus
  0.5em minus 0.4em\relax International Society for Optics and Photonics, 2001,
  pp. 57--65.

\bibitem{ICANN18}
I.~Grenet, Y.~Yin, J.-P. Comet, and E.~Gelenbe, ``Machine learning to predict
  toxicity of compounds,'' in \emph{27th Annual International Conference on
  Artificial Neural Networks, ICANN18, accepted for publication}.\hskip 1em
  plus 0.5em minus 0.4em\relax Springer Verlang, 2018.

\bibitem{CPN}
E.~Gelenbe, ``Steps toward self-aware networks,'' \emph{Communications of the
  ACM}, vol.~52, no.~7, pp. 66--75, 2009.

\bibitem{Zarina}
E.~Gelenbe and Z.~Kazhmaganbetova, ``Cognitive packet network for bilateral
  asymmetric connections,'' \emph{IEEE Trans. Industrial Informatics}, vol.~10,
  no.~3, pp. 1717--1725, 2014.

\bibitem{Brun}
O.~Brun, L.~Wang, and E.~Gelenbe, ``Big data for autonomic intercontinental
  overlays,'' \emph{IEEE Journal on Selected Areas in Communications}, vol.~34,
  no.~3, pp. 575--583, 2016.

\bibitem{Francois1}
F.~Fran\c{c}ois and E.~Gelenbe, ``Towards a cognitive routing engine for
  software defined networks,'' in \emph{ICC 2016}.\hskip 1em plus 0.5em minus
  0.4em\relax IEEE, 2016, pp. 1--6.

\bibitem{Francois2}
------, ``Optimizing secure sdn-enabled inter-data centre overlay networks
  through cognitive routing,'' in \emph{MASCOTS 2016, IEEE Computer
  Society}.\hskip 1em plus 0.5em minus 0.4em\relax IEEE, 2016, pp. 283--288.

\bibitem{Wang2}
L.~Wang and E.~Gelenbe, ``Adaptive dispatching of tasks in the cloud,''
  \emph{IEEE Transactions on Cloud Computing}, vol.~6, no.~1, pp. 33--45, 2018.

\bibitem{Worms}
\BIBentryALTinterwordspacing
G.~Sakellari and E.~Gelenbe, ``Demonstrating cognitive packet network
  resilience to worm attacks,'' in \emph{Proceedings of the 17th {ACM}
  Conference on Computer and Communications Security, {CCS} 2010, Chicago,
  Illinois, USA, October 4-8, 2010}, 2010, pp. 636--638. [Online]. Available:
  \url{http://doi.acm.org/10.1145/1866307.1866380}
\BIBentrySTDinterwordspacing

\bibitem{IoT}
O.~Brun, Y.~Yin, E.~Gelenbe, Y.~M. Kadioglu, J.~Augusto-Gonzalez, and M.~Ramos,
  ``Deep learning with dense random neural networks for detecting attacks
  against iot-connected home environments,'' in \emph{Security in Computer and
  Information Sciences: First International ISCIS Security Workshop 2018,
  Euro-CYBERSEC 2018, London, UK, February 26-27, 2018}.\hskip 1em plus 0.5em
  minus 0.4em\relax Lecture Notes CCIS No. 821, Springer Verlag, 2018.

\bibitem{GN1}
E.~Gelenbe, ``Product-form queueing networks with negative and positive
  customers,'' \emph{Journal of applied probability}, vol.~28, no.~3, pp.
  656--663, 1991.

\bibitem{GN3}
E.~Gelenbe and R.~Schassberger, ``Stability of product form g-networks,''
  \emph{Probability in the Engineering and Informational Sciences}, vol.~6,
  no.~3, pp. 271--276, 1992.

\bibitem{GN2}
E.~Gelenbe, P.~Glynn, and K.~Sigman, ``Queues with negative arrivals,''
  \emph{Journal of applied probability}, vol.~28, no.~1, pp. 245--250, 1991.

\bibitem{GN4}
E.~Gelenbe, ``G-networks by triggered customer movement,'' \emph{Journal of
  applied probability}, vol.~30, no.~3, pp. 742--748, 1993.

\bibitem{GN5}
------, ``G-networks with signals and batch removal,'' \emph{Probability in the
  Engineering and Informational Sciences}, vol.~7, no.~3, pp. 335--342, 1993.

\bibitem{GN6}
J.-M. Fourneau and E.~Gelenbe, ``G-networks with adders,'' \emph{Future
  Internet}, vol.~9, no.~3, p.~34, 2017.

\bibitem{Morfo}
E.~Gelenbe and C.~Morfopoulou, ``A framework for energy-aware routing in packet
  networks,'' \emph{The Computer Journal}, vol.~54, no.~6, p. 850–859, 2010.

\bibitem{GelenbeRegNets2007}
E.~Gelenbe, ``Steady-state solution of probabilistic gene regulatory
  networks,'' \emph{Physical Review E}, vol.~76, 2007.

\bibitem{KimG12}
\BIBentryALTinterwordspacing
H.~Kim and E.~Gelenbe, ``Stochastic gene expression modeling with hill function
  for switch-like gene responses,'' \emph{{IEEE/ACM} Trans. Comput. Biology
  Bioinform.}, vol.~9, no.~4, pp. 973--979, 2012. [Online]. Available:
  \url{https://doi.org/10.1109/TCBB.2011.153}
\BIBentrySTDinterwordspacing

\bibitem{EPN}
E.~Gelenbe, ``Energy packet networks: Adaptive energy management for the
  cloud,'' in \emph{CloudCP '12 Proceedings of the 2nd International Workshop
  on Cloud Computing Platforms}.\hskip 1em plus 0.5em minus 0.4em\relax ACM,
  2012, p.~1.

\bibitem{Marin}
E.~Gelenbe and A.~Marin, ``Interconnected wireless sensors with energy
  harvesting,'' in \emph{Analytical and Stochastic Modelling Techniques and
  Applications - 22nd International Conference, {ASMTA} 2015, Albena, Bulgaria,
  May 26-29, 2015. Proceedings}, 2015, pp. 87--99.

\bibitem{Fourneau}
J.~Fourneau, A.~Marin, and S.~Balsamo, ``Modeling energy packets networks in
  the presence of failures,'' in \emph{24th {IEEE} International Symposium on
  Modeling, Analysis and Simulation of Computer and Telecommunication Systems,
  {MASCOTS} 2016, London, United Kingdom, September 19-21, 2016}, 2016, pp.
  144--153.

\bibitem{Ceran2}
E.~Gelenbe and E.~T. Ceran, ``Energy packet networks with energy harvesting,''
  \emph{IEEE Access}, vol.~4, p. 1321–1331, 2016.

\bibitem{NOLTA}
E.~Gelenbe and O.~H. Abdelrahman, ``An energy packet network model for mobile
  networks with energy harvesting,'' \emph{Nonlinear Theory and Its
  Applications, IEICE}, vol.~9, no.~3, p. 1–15, 2018.

\bibitem{gelenbe2006function}
E.~Gelenbe, Z.-H. Mao, and Y.-D. Li, ``Function approximation by random neural
  networks with a bounded number of layers,'' in \emph{Computer System
  Performance Modeling In Perspective: A Tribute to the Work of Prof Kenneth C
  Sevcik}.\hskip 1em plus 0.5em minus 0.4em\relax World Scientific, 2006, pp.
  35--58.

\bibitem{gelenbe1999function}
------, ``Function approximation with spiked random networks,'' \emph{IEEE
  Transactions on Neural Networks}, vol.~10, no.~1, pp. 3--9, 1999.

\bibitem{hornik1989multilayer}
K.~Hornik, M.~Stinchcombe, and H.~White, ``Multilayer feedforward networks are
  universal approximators,'' \emph{Neural networks}, vol.~2, no.~5, pp.
  359--366, 1989.

\bibitem{Multiclass}
E.~Gelenbe and K.~F. Hussain, ``Learning in the multiple class random neural
  network,'' \emph{IEEE Transactions on Neural Networks}, vol.~13, no.~6, pp.
  1257--1267, 2002.

\bibitem{gelenbe2001learning}
------, ``Learning and generating color textures with recurrent multiple class
  random neural networks,'' in \emph{Applications of Artificial Neural Networks
  in Image Processing VI}, vol. 4305.\hskip 1em plus 0.5em minus 0.4em\relax
  International Society for Optics and Photonics, 2001, pp. 49--57.

\bibitem{Laser}
H.~M. Abdelbaki, K.~Hussain, and E.~Gelenbe, ``A laser intensity image based
  automatic vehicle classification system,'' in \emph{Intelligent
  Transportation Systems, 2001. Proceedings}.\hskip 1em plus 0.5em minus
  0.4em\relax IEEE, 2001, pp. 460--465.

\bibitem{Simulation}
E.~Gelenbe, K.~Hussain, and V.~Kaptan, ``Simulating autonomous agents in
  augmented reality,'' \emph{Journal of Systems and Software}, vol.~74, no.~3,
  pp. 255--268, 2005.

\bibitem{lichman2013uci}
M.~Lichman \emph{et~al.}, ``Uci machine learning repository,'' 2013.

\bibitem{petricoin2002use}
E.~F. Petricoin~III, A.~M. Ardekani, B.~A. Hitt, P.~J. Levine, V.~A. Fusaro,
  S.~M. Steinberg, G.~B. Mills, C.~Simone, D.~A. Fishman, E.~C. Kohn
  \emph{et~al.}, ``Use of proteomic patterns in serum to identify ovarian
  cancer,'' \emph{The lancet}, vol. 359, no. 9306, pp. 572--577, 2002.

\bibitem{hussain2016road}
K.~F. Hussain and G.~S. Moussa, ``On-road vehicle classification based on
  random neural network and bag-of-visual words,'' \emph{Probability in the
  Engineering and Informational Sciences}, vol.~30, no.~3, pp. 403--412, 2016.

\bibitem{Learning93}
E.~Gelenbe, ``Learning in the recurrent random neural network,'' \emph{Neural
  Computation}, vol.~5, no.~1, pp. 154--164, 1993.

\bibitem{Learning02}
E.~Gelenbe and K.~F. Hussain, ``Learning in the multiple class random neural
  network,'' \emph{IEEE Transactions on Neural Networks}, vol.~13, no.~6, pp.
  1257--1267, 2002.

\bibitem{fernandez2014we}
M.~Fern{\'a}ndez-Delgado, E.~Cernadas, S.~Barro, and D.~Amorim, ``Do we need
  hundreds of classifiers to solve real world classification problems?''
  \emph{The Journal of Machine Learning Research}, vol.~15, no.~1, pp.
  3133--3181, 2014.

\bibitem{gelenbe1989random}
E.~Gelenbe, ``Random neural networks with negative and positive signals and
  product form solution,'' \emph{Neural computation}, vol.~1, no.~4, pp.
  502--510, 1989.

\end{thebibliography}

%%%%%%%%%%%%%%%%%%%%%%%%%%%%%%%%%%%%%%%%%%

\end{document}